\documentclass[11pt,a4paper]{article}

\usepackage[a4paper,margin=2.5cm]{geometry}
\usepackage[T1]{fontenc}
\usepackage[utf8]{inputenc}
\usepackage{libertinus}
\usepackage{microtype}
\usepackage{amsmath,amssymb}
\usepackage{graphicx}
\usepackage{booktabs}
\usepackage{array}
\usepackage{tabularx}
\usepackage{siunitx}
\usepackage{xcolor}
\usepackage{csquotes}
\usepackage{algorithm}
\usepackage{algpseudocode}
\usepackage[numbers]{natbib}
\usepackage[hidelinks]{hyperref}
\usepackage[capitalise,noabbrev]{cleveref}

\newenvironment{keywords}{\par\medskip\noindent\textbf{Keywords:}\ }{\par}
\newcommand{\sep}{\unskip, }

\title{\textbf{From Certified Pairs to a Certified Hub: Reasoner-Checked
  Crosswalks for the Information Exchange Standard}}
\author{Fabio Rovai\\[0.3em]
  \normalsize The Tesseract Academy (Kampakis and Co Ltd), London, United Kingdom\\
  \normalsize\href{mailto:fabio@thetesseractacademy.com}{\texttt{fabio@thetesseractacademy.com}}
  \quad ORCID \href{https://orcid.org/0009-0001-7244-794X}{0009-0001-7244-794X}}
\date{Preprint, November 2026}

\begin{document}
\maketitle

\begin{abstract}
A national digital twin was conceived as an ecosystem of twins connected through
mappings to a common foundation, but the roadmap that promises to confirm those
mappings are valid never says what valid means. We give the test a precise
form, coherence and conservativity of the merged ontology, borrowed from
logic-based alignment repair, and apply it to the Information Exchange Standard
(IES), first pair by pair and then to whole hubs. Crosswalks from IES into
thirteen ontologies across seven sector groups, anchored on the building
ontologies BOT, Brick, and SAREF4BLDG, pass through separate stages of matching,
human review, certification of each SKOS correspondence against the two class
hierarchies, and reasoning with HermiT. Pair by pair, the stages catch different
failures: certification alone removed all \num{526} conservativity violations of
the raw crosswalks, review alone left \num{73}, and together they made eleven of
twelve overlapping pairs coherent and conservative. Pairwise certification does
not compose. Merging the three certified building crosswalks into one hub made
five Brick classes unsatisfiable, through a homonym between the IES and Brick
notions of a floor that only BOT's disjointness axioms expose; rejecting that
one row gave a coherent and conservative hub that entails \num{290}
cross-ontology links nobody wrote down. The core hub is inconsistent before any
crosswalk is added, because two published standards, the Open Energy Ontology
and BFO~2020, contradict each other; with either one left out it is coherent
and conservative. A hub, not a pair, is the unit an ecosystem of twins has to
certify.
\end{abstract}

\begin{keywords}
  ontology matching \sep
  alignment coherence \sep
  conservativity \sep
  crosswalks \sep
  Information Exchange Standard \sep
  building ontologies \sep
  digital twins
\end{keywords}

\section{Introduction}
\label{sec:intro}

The UK's national digital twin was never meant to be one model. Its founding
principles define it as ``an ecosystem of digital twins connected via securely
shared data'' \cite{gemini2018}, and the roadmap that followed rests that
ecosystem on mappings: existing models and data in other vocabularies are to be
connected to a common foundation, and ``permanent automated mappings will allow
the digital twin catalogue to be continuously refreshed''
\cite{hetherington2020imf}. The same roadmap promises to reload the source twins
``to confirm that the new mappings are valid'' \cite{hetherington2020imf}, but
it does not say what makes a mapping valid. This paper proposes a test that
does, applies it to every mapping and to every hub of mappings we could build
around one common ontology, and reports where the ecosystem holds and where it
breaks.

Such an ecosystem is only as useful as the data that flows into it, and
that data arrives in whatever vocabulary its source happens to use. Building
records speak the language of the construction and facilities-management
community, energy records speak the language of grid models, and sensor feeds
speak the language of observation ontologies. The Information Exchange Standard
(IES) was designed as a common 4D ontology for exchanging information across
government, with a lineage in high-quality 4D data modelling and formal
upper-ontology work \cite{ies4,west2011hqdm}. It gives a shared backbone, but
it does not, on its own, tell a data engineer how an IES class relates to the
corresponding class in a building or energy ontology. That relationship has to
be established by an alignment, and today those alignments are written by hand,
reviewed informally, and shipped without a formal test that the result is even
logically coherent.

The cost of this shows up whenever a twin has to grow. Extending the vocabulary
to cover a new asset class, or connecting a new domain data source, means
writing and checking new mappings by hand, and there is no agreed test that
says when such a step is safe. The
research community has strong machinery for each half of the problem
separately. Ontology matching is a mature field with an established evaluation
tradition \cite{euzenat2013ontologymatching}, SKOS provides a vocabulary for
typing correspondences \cite{miles2009skos}, SSSOM provides a way to serialize
and share them \cite{matentzoglu2022sssom}, and OWL~2 reasoners such as HermiT
can decide the consistency of the merged result \cite{glimm2014hermit}. Using
the reasoner to judge an alignment is also established: logic-based matchers
repair incoherent mappings \cite{jimenezruiz2011logmap}, reasoning support
helps experts revise them \cite{meilicke2009revision}, and the conservativity
principle flags mappings that change the meaning of an input's own terms
\cite{solimando2014conservativity}. Pointed at IES, this machinery has so far
produced only unreviewed output: a recent bulk run of LogMap over defence and
national-security ontologies aligned IES with several building ontologies,
among many other pairs, but those alignments were not reviewed and were not
tested for conservativity \cite{dilworth2026defence}. We are not aware of a
public crosswalk that takes IES to the domain ontologies a twin consumes,
reviews and types each correspondence, tests it against both class
hierarchies, and reports what the reasoner makes of the merge.

In earlier work we aligned three upper ontologies that sit under UK and allied
defence data infrastructure, IES, HQDM, and BFO, and introduced an
orientation-search procedure that produces a reasoner-checked crosswalk rather
than a hand-asserted one \cite{rovai2026coherence}. That work also recorded a
plain fact about the state of the art: no public alignment between IES and HQDM
existed before it. The present paper takes the acceptance test of that work,
coherence and conservativity of the merge, off that single upper-ontology
triangle and applies it across the domain ontologies a working twin actually
consumes, using a lighter pipeline that runs unchanged on every pair.
The underlying alignment engine is open source and has been evaluated on
standard matching benchmarks \cite{rovai2026openontologies}; here we are
concerned with the step that follows matching, namely certifying that a
proposed crosswalk is safe to merge.

We anchor the paper on the building domain for three reasons. Building data is
plentiful and open, through public building-performance certificates
\cite{mhclg_epc_data}, the
building-ontology landscape is deep and well studied, and extending an IES building vocabulary to regional or national
retrofit scale is exactly the kind of step that should be small and checkable
rather than a project of its own. That landscape is itself prior art. Around BOT
\cite{rasmussen2021bot}, the core vocabulary of the W3C Linked Building Data
community, building ontologies have been aligned for years: BOT ships alignment
modules to Brick \cite{balaji2018brick}, ifcOWL \cite{pauwels2016ifcowl}, and
SAREF4BLDG \cite{etsi2020saref4bldg}, among others \cite{bot2021spec}.
We do not claim to have aligned building ontologies for the first time. Our
claim is narrower and specific to the IES side: we are not aware of prior
published certified alignments between IES and this established
building-ontology landscape, where certified means reasoner-checked for
coherence and conservativity rather than agreed by consensus or validated by
hand.

This paper makes four contributions, each defended in a later section.

First, we set out a certified IES-to-domain crosswalk pipeline in which
matching, human review, and certification are separate stages: SKOS-typed
correspondences are downgraded when the class hierarchies cannot support the
claimed predicate, and a crosswalk is accepted only when an OWL~2 reasoner finds
the merge coherent and conservative over each input's own terms
(\cref{sec:method}).

Second, we apply the pipeline to the building-ontology landscape as an anchor
case and lay out the full crosswalk matrix for a cross-sector IES hub, spanning
energy, geospatial, sensor, provenance, temporal, catalogue, and foundation
ontologies, with the licensing constraints of each source recorded so that
copyrighted standards are referenced for alignment and never redistributed
(\cref{sec:method,sec:eval}).

Third, we measure each pair under four crosswalks, raw, certified without
review, reviewed, and reviewed then certified, which separates what each stage
contributes. Certification alone removes every conservativity violation but no
unsatisfiability; review alone removes the homonyms but leaves violations; together
they make eleven of twelve overlapping pairs coherent and conservative. The
twelfth, HQDM, stays incoherent, and we trace that to a single correspondence onto
a class HQDM cannot populate, a failure the hierarchy test cannot see
(\cref{sec:eval}).

Fourth, we run the same test on whole hubs, every target of one IES profile and
every certified crosswalk merged at once, and show that pairwise certification
does not compose: a homonym invisible in each pair breaks the building hub, and
two published standards make the core hub inconsistent before any crosswalk is
added. Once repaired, the hubs are coherent and conservative and entail
cross-ontology links that no one stated (\cref{sec:eval-hub}).

We position this work against the closest prior effort, a recent evaluation
resource for defence and national-security ontology matching that builds a
consensus from several matchers, validates it by hand into a silver standard,
and measures the incoherence each alignment induces \cite{dilworth2026defence}.
IES is in that resource's collection but in none of its matching tasks. That
resource scores matchers against a reference; this paper produces usable IES
crosswalks with their measured coherence and conservativity, which could in turn
seed an IES task. We return to this contrast in \cref{sec:related}.

\section{Background and Related Work}
\label{sec:related}

\subsection{Ontology matching and its evaluation}

Ontology matching produces correspondences between the entities of two
ontologies, and the field has a long-standing body of techniques and a shared
evaluation tradition \cite{euzenat2013ontologymatching}. The usual output of a
matcher is a set of candidate correspondences, each with a confidence score and
a relation such as equivalence or subsumption. The question this paper is
concerned with begins where matching ends. A high-confidence candidate is a
claim about meaning, and merging the two ontologies under that claim can make
the combined ontology say things neither author intended, or make it
inconsistent outright. Matching confidence does not measure that risk, so we
treat the matcher as one stage and the logical acceptance test as a separate
stage.

\subsection{Logic-based alignment repair}
\label{sec:related-repair}

Using a reasoner to judge an alignment is itself well established, and our
acceptance test borrows its two criteria from this line of work. Reasoning-based
mapping revision treats an alignment as incoherent when the two ontologies
together with the mappings make some class unsatisfiable, and uses the reasoner
to propagate the logical consequences of an expert's decision to accept or
reject a mapping \cite{meilicke2009revision}. LogMap
builds that check into a scalable matcher, repairing its own candidate mappings
as it goes \cite{jimenezruiz2011logmap}. The conservativity principle adds a
second, weaker failure: an alignment should not introduce new subsumptions
between the terms of a single input ontology, even when every class stays
satisfiable, and such violations can be detected and corrected
automatically \cite{solimando2014conservativity,solimando2017minimizing}. Our
coherence test is the first criterion and our conservativity count is the
second. What we add is not the criteria but their application, with a
SKOS-strength downgrade and human review, to IES and the domain ontologies
around it, where the only public alignments so far are unreviewed matcher
output (\cref{sec:related-closest}).

\subsection{The closest prior work}
\label{sec:related-closest}

The nearest effort to ours is a recent evaluation resource that improves
interoperability among defence and national-security ontologies
\cite{dilworth2026defence}. It analyses more than sixty public ontologies from
that space and introduces a new track, with eight matching tasks, for the
Ontology Alignment Evaluation Initiative. Its consensus alignments are, in the
authors' words, derived by aggregating the outputs of several state-of-the-art
ontology alignment systems, and its silver standard is obtained from the manual
validation of the consensus alignment together with a subset of mappings that
only one system proposed. The resource takes coherence seriously: it counts the
unsatisfiable classes that each system, each consensus, and the silver standard
induce in the merged ontology, finds that even manually curated mappings can be
incoherent, and treats the silver standard to fix those logical errors.

Two things separate the present work from it. The first is coverage. IES is in
that resource's ontology collection, but none of its eight matching tasks
involves IES, and the authors note that a complete and formal alignment between
IES and BFO still needs to be established. The tasks fall into three
subdomains, cybersecurity, situation awareness, and smart environments, and four
of the eight involve Brick. Before choosing the tasks, the authors ran LogMap
over 1,653 ontology pairs of the collection and published the output, which
includes IES against BOT, Brick, and SAREF4BLDG. That output proposes, among
others, the same \texttt{Element} homonym between IES and BOT that we report in
\cref{sec:eval-coherence}; it was used to select tasks and was not reviewed or
tested for conservativity. The second is the form of the output. A silver
standard is a reference against which systems are scored; what we produce is a
crosswalk meant to be used, in which each correspondence carries the SKOS
strength that the class hierarchies support and each pair carries its measured
coherence and conservativity. The two fit together. An IES task could be added
to such a track, and the crosswalks here are candidates for it, to be scored
against whatever reference that community builds. We present this work as
complementary to that resource and not as a correction of it.

\subsection{The building-ontology landscape as prior art}
\label{sec:related-building}

The building domain already has a rich and interrelated set of ontologies, and
aligning them is not new. The Building Topology Ontology (BOT) is the core
vocabulary of the W3C Linked Building Data community group, giving a high-level
description of the topology of buildings including storeys and spaces, the
building elements they contain, and their 3D models \cite{rasmussen2021bot}.
Brick is a metadata schema for the sensors, subsystems, and relationships that
drive portable building applications \cite{balaji2018brick}. The ifcOWL
ontology renders the Industry Foundation Classes model in OWL for the
construction industry \cite{pauwels2016ifcowl}. SAREF, the Smart Applications
REFerence ontology, and its building extension SAREF4BLDG are published as ETSI
Technical Specifications \cite{saref,etsi2020saref4bldg}. BOT, as the central
vocabulary of the Linked Building Data effort, ships alignment modules to
Brick, ifcOWL, SAREF4BLDG, and several other building ontologies
\cite{bot2021spec}. We build on this landscape and cite it as the established starting
point; the novelty we claim is only on the IES side, as set out in
\cref{sec:intro}.

\subsection{IES and the upper-ontology layer}
\label{sec:related-ies}

IES is a 4D ontology serialized as an RDF Schema, developed for information
exchange across government and expressed over a small set of foundational
notions such as entities, states, and the events that relate them
\cite{ies4}. Its intellectual lineage runs through high-quality 4D data
modelling \cite{west2011hqdm} and connects to the wider top-level ontology
discussion, including the Basic Formal Ontology (BFO), which is standardized
internationally \cite{arp2015bfo,isoiec21838}. Bailey et al.\ compared IES and BFO
design patterns and proposed high-level mapping patterns between them, leaving
axiomatic translation to future work \cite{bailey2025iesbfo}. Our earlier paper
aligned IES, HQDM, and BFO axiomatically, reported that no public alignment
between IES and HQDM existed before it, and certified the result with a
reasoner \cite{rovai2026coherence}. That paper is the base this one extends
outward toward the domain ontologies.

\subsection{Mapping vocabularies and reasoning}
\label{sec:related-mapping}

Two standards carry the correspondences through the pipeline. SKOS supplies the
mapping predicates, a common data model for sharing and linking knowledge
organization systems via the Web, with relations for exact, broader, narrower,
close, and related matches \cite{miles2009skos}. SSSOM provides a tabular
serialization in which each correspondence carries its predicate and its
provenance, so that a crosswalk is a reviewable artifact rather than an opaque
blob \cite{matentzoglu2022sssom}. The acceptance test is a reasoning problem:
whether each ontology is satisfiable alone, whether the merged ontology is
satisfiable, and whether the merge is conservative in the sense of forcing no
new subclass relationships onto either input's own vocabulary. We use the
HermiT OWL~2 reasoner for these checks \cite{glimm2014hermit}. Beyond the
building anchor, the crosswalk matrix draws on domain standards that are already
established in their fields: SOSA and SSN for sensors and observations
\cite{janowicz2019sosa,haller2019ssn}, GeoSPARQL for geospatial data
\cite{ogc2024geosparql,battle2012geosparql}, PROV-O for provenance \cite{provo}, the Time
Ontology for temporal structure \cite{owltime}, DCAT for data catalogues
\cite{dcat3}, and public domain ontologies such as the beAWARE
emergency-management ontology \cite{beaware}.

\section{Method}
\label{sec:method}

\subsection{Overview}

The pipeline takes two ontologies, a source and a target, and produces a
crosswalk together with the evidence for accepting it. It runs in five stages:
align, review, certify, reason, and serialize. Each stage writes an artifact to
disk, so a crosswalk can be inspected, re-run, and reproduced from the command
line. The design keeps the proposal of a correspondence apart from its
acceptance. The matcher proposes, a reviewer removes homonyms and adds
correspondences the matcher could not see, the certifier sets the strength of
each correspondence to what the two class hierarchies support, and the reasoner
checks the merge. A correspondence whose claimed relation is too strong is
weakened rather than discarded, so that information is kept where the evidence
supports a weaker claim. Because review and certification are separate stages,
we can measure what each contributes on its own (\cref{sec:eval-coherence}).

\subsection{Candidate alignment}
\label{sec:method-align}

The first stage uses the open-source Open Ontologies alignment engine, whose
evaluation found stable one-to-one matching to be the dominant factor in
alignment quality, and which has been evaluated on standard benchmarks
\cite{rovai2026openontologies}. We run it in a deterministic configuration: a
single confidence threshold, one-to-one cardinality, and an informative scoring
scheme in which a candidate's confidence is the weighted mean of only those
signals that both sides carry data for, rather than counting an unmeasurable
signal as zero. Each alignment is run three times and the three candidate sets,
each a set of subject, object, relation, and confidence, are compared for
equality, so that any reported timing is a median over repeats and the result
is confirmed to be reproducible rather than stochastic. Candidates that are ambiguous or that fall on the threshold are held
back for review rather than applied silently.

This paper is not about raising matcher recall, and the stage is deliberately
lighter than the orientation search of our earlier upper-ontology work
\cite{rovai2026coherence}. Orientation search treats the direction of each
correspondence as a search variable and uses the reasoner as an oracle while it
searches; here the reasoner is used only once, as a final check on a crosswalk
that matching, review, and certification have already fixed. That is cheaper and
runs unchanged on every pair, and \cref{sec:eval-coherence} shows where it is
enough and where it is not. A low candidate count against a large source
vocabulary is expected and is not treated as a failure: most classes in a large
upper or domain ontology have no counterpart in a given target, and the
pipeline records how many source classes remain unmapped rather than inventing
correspondences to cover them.

\subsection{Typing correspondences with SKOS and SSSOM}
\label{sec:method-type}

Each surviving candidate is assigned a SKOS mapping predicate
\cite{miles2009skos}. An engine equivalence becomes \texttt{skos:exactMatch},
a subsumption becomes \texttt{skos:broadMatch} or \texttt{skos:narrowMatch}
according to direction, and weaker associations become \texttt{skos:closeMatch}
or \texttt{skos:relatedMatch}. The typed correspondences are written as an SSSOM
table \cite{matentzoglu2022sssom}, one row per correspondence, each row carrying
its subject, object, predicate, and the origin of the row, whether proposed by
the engine or added during curation. Curation is a recorded decision per engine
row, to keep, revise, or reject it, with a one-line reason read against the two
standards' own definitions, plus any correspondence the reviewer adds. The most
common rejection is a homonym, a pair of classes with the same name and a
different meaning, such as a device's on or off state against an IES state,
which is a temporal part of an individual. This serialization is what a domain expert
reviews and what downstream tools consume, and it is the form in which a
crosswalk is published.

\subsection{Crosswalk certification}
\label{sec:method-certify}

Certification tests each typed correspondence against the class hierarchies of
the two ontologies (\cref{alg:certify}). Each ontology is materialised on its
own under the engine's OWL~2 RL rules, and the ancestors of a class are its
entailed named superclasses, together with its broader concepts when the input
is a SKOS scheme. The crosswalk is then used as its own translation table: the
subject's ancestors are carried into the target vocabulary through the rows
that map them, and compared with those of the object's ancestors that some row
maps onto. The question is whether the two translated ancestor sets support the
claimed predicate. The test has three possible outcomes for a row.

A row is \emph{downgraded} when the hierarchy supports a weaker predicate than
the one claimed. The characteristic case is an \texttt{exactMatch} whose object
turns out to sit under every ancestor the subject sits under, and more, so the
object is the more specific concept and the supported relation is a
\texttt{narrowMatch}. Downgrading keeps the correspondence while correcting its
strength, which is safer than either accepting an overstated equivalence or
throwing the pair away.

A row is \emph{upheld with a coverage gap} when both sides entail the same
ancestors over every ancestor the mapping lets them compare, but some ancestors
on one or both sides have no counterpart through the mapping and so could not be
compared. The uncompared ancestors are a gap in the mapping's coverage, not
evidence that the two terms differ, and the pipeline records exactly which
ancestors were untranslatable so that the gap is visible.

A row is \emph{not refuted, nothing compared} when neither term has an ancestor
that the mapping carries across, so the hierarchy can neither support nor refute
the claim. Such a row is explicitly not certified, and the output says so rather
than letting a bare label-match pass as if it had been checked. The pipeline
reports the counts in each bucket, the number of rows downgraded, and the number
of source classes left unmapped, so that the strength of a crosswalk can be read
off directly.

\begin{algorithm}[t]
\caption{Certifying the strength of each correspondence}
\label{alg:certify}
\begin{algorithmic}[1]
\Require source $O_S$, target $O_T$, crosswalk $M$ of rows $(s, p, o)$ with $p$ a SKOS predicate
\Ensure each row with a written predicate and an outcome; the unmapped source classes
\State $A_S \gets$ entailed named superclasses in $O_S$ under OWL~2 RL, plus \texttt{skos:broader} closure
\State $A_T \gets$ the same for $O_T$
\State $\tau \gets \{ s \mapsto o : (s, p, o) \in M \}$ \Comment{the crosswalk is its own translation table}
\State $I \gets \{ o : (s, p, o) \in M \}$ \Comment{target terms some row maps onto}
\ForAll{$(s, p, o) \in M$}
  \State $X \gets \{ \tau(a) : a \in A_S(s) \cap \mathrm{dom}\,\tau \}$; \quad $U \gets A_S(s) \setminus \mathrm{dom}\,\tau$
  \State $Y \gets A_T(o) \cap I$; \quad $V \gets A_T(o) \setminus I$ \Comment{$U, V$: coverage gaps}
  \State $q \gets$ \textsf{exact} if $X = Y$; \textsf{broad} if $X \supset Y$; \textsf{narrow} if $Y \supset X$; else \textsf{related}
  \If{$p$ is stronger than $q$ in the order \textsf{exact} $>$ \textsf{broad}, \textsf{narrow} $>$ \textsf{related}}
    \State write $q$; outcome \emph{downgraded}
  \ElsIf{$X = \emptyset$ \textbf{and} $Y = \emptyset$}
    \State write $p$; outcome \emph{not refuted, nothing compared}
  \ElsIf{$U \neq \emptyset$ \textbf{or} $V \neq \emptyset$}
    \State write $p$; outcome \emph{upheld with a coverage gap}
  \Else
    \State write $p$; outcome \emph{upheld}
  \EndIf
\EndFor
\State \Return rows, and every class of $O_S$ with ancestors and no row, as unmapped
\end{algorithmic}
\end{algorithm}

Two properties of \cref{alg:certify} matter for what follows. It never
strengthens a claim: a row is only ever rewritten to a weaker predicate, and a
\texttt{closeMatch}, which sits outside the order, is never rewritten at all,
because two sibling classes under one mapped parent have identical translated
ancestors and a compatible hierarchy is not evidence of identity. And it reads
only subsumption. It does not see disjointness axioms, and it does not know
whether a class is satisfiable in its own ontology, since materialising a
schema with no individuals derives no clash. Both blind spots matter, and both
are why the reasoning stage below is a separate test rather than a formality.

\subsection{Reasoning: coherence and conservativity}
\label{sec:method-reason}

The final acceptance test promotes the SKOS crosswalk into OWL axioms and
reasons over the merge with HermiT \cite{glimm2014hermit}. The promotion is
fixed: \texttt{exactMatch} becomes an \texttt{owl:equivalentClass} bridge
axiom, \texttt{broadMatch} becomes the subject being a subclass of the object,
\texttt{narrowMatch} becomes the object being a subclass of the subject, and
\texttt{closeMatch} and \texttt{relatedMatch} produce no axiom because they
assert no subsumption. The reasoner is first run on the source alone and on the
target alone, to record which classes each input already cannot populate. It is
then run on the merge under four crosswalks: the raw engine output, the raw
output after certification with no human review, the reviewed crosswalk, and
the reviewed crosswalk after certification. The four conditions separate what
review and certification each contribute. For every merge the reasoner reports
how many bridge axioms were added and which classes became unsatisfiable that
were satisfiable in their own ontology.

Coherence is necessary but not sufficient, because a merge can keep every class
satisfiable while still forcing new subclass relationships onto one input's own
vocabulary, a side effect that would silently change the meaning of a standard.
Following the general notion of the conservativity principle
\cite{solimando2014conservativity}, without the exemption their variant grants
to classes that already share a descendant, the pipeline therefore also counts violations: subclass relationships between two
named, satisfiable classes of a single input that the merge entails and that
input alone does not. A crosswalk is reported as certified when the merge is
consistent, no class is newly unsatisfiable, and the conservativity violation
count is zero on both sides. The condition is stated relative to each input,
not as a requirement that each input be coherent, because one published upper
ontology in our matrix already has unsatisfiable classes of its own
(\cref{sec:eval-coherence}). Counting only subsumptions between named classes
makes the conservativity test an approximation: a positive count is a real
change of meaning, while a zero count rules out only this kind of change.
\Cref{alg:accept} sets the test out in full.

\begin{algorithm}[t]
\caption{Acceptance test for one crosswalk}
\label{alg:accept}
\begin{algorithmic}[1]
\Require source $O_S$, target $O_T$ with named classes $\Sigma_S, \Sigma_T$; SKOS crosswalk $M$
\Ensure accept or reject, with the newly unsatisfiable classes $N$ and the violations $V_S, V_T$
\State $B \gets \{ s \equiv o : (s, \textsf{exact}, o) \in M \} \cup \{ s \sqsubseteq o : (s, \textsf{broad}, o) \in M \} \cup \{ o \sqsubseteq s : (s, \textsf{narrow}, o) \in M \}$
\Statex \hfill\Comment{\textsf{close} and \textsf{related} rows add no axiom}
\State $Z_S \gets \textsc{Unsat}(O_S)$; \quad $Z_T \gets \textsc{Unsat}(O_T)$ \Comment{HermiT, each input alone}
\State $H_S \gets \textsc{Classify}(O_S)^{+}$; \quad $H_T \gets \textsc{Classify}(O_T)^{+}$ \Comment{transitive closure, named classes}
\State $O \gets O_S \cup O_T \cup B$ \Comment{imports dropped: nothing is fetched}
\If{$O$ is inconsistent} \State \Return reject \EndIf
\State $Z \gets \textsc{Unsat}(O)$; \quad $N \gets Z \setminus (Z_S \cup Z_T)$
\State $H \gets \textsc{Classify}(O)^{+}$
\ForAll{$X \in \{S, T\}$}
  \State $V_X \gets \{ (a, b) : a, b \in \Sigma_X,\ a \notin Z,\ a \neq b,\ a \sqsubseteq_H b,\ a \not\sqsubseteq_{H_X} b \}$
\EndFor
\State \Return accept iff $N = \emptyset$ and $V_S = V_T = \emptyset$
\end{algorithmic}
\end{algorithm}

The evaluation runs \cref{alg:accept} on each pair four times, with $M$ the raw
engine crosswalk, the raw crosswalk after \cref{alg:certify}, the reviewed
crosswalk, and the reviewed crosswalk after \cref{alg:certify}.

\subsection{The crosswalk matrix and the building anchor}
\label{sec:method-matrix}

We take the building domain as the anchor case because it combines plentiful
open data with a deep target landscape. The building targets we align to are
BOT \cite{rasmussen2021bot}, Brick \cite{balaji2018brick}, and SAREF4BLDG
\cite{etsi2020saref4bldg}; ifcOWL \cite{pauwels2016ifcowl} belongs to the same landscape but
was not obtained as an openly fetchable OWL file in this offline run
(\cref{sec:eval-skip}). The empirical input
is open building-performance certificate data \cite{mhclg_epc_data}, from which an IES building
vocabulary can be extended for regional retrofit and the extension measured for
size and reasoning cost. Around this anchor sits a wider matrix intended to show
that a single certified IES hub can reach across sectors: energy, through
openly licensed energy vocabularies and, by reference only, the copyrighted grid
model standards; infrastructure and assets, through GeoSPARQL
\cite{ogc2024geosparql,battle2012geosparql}; incident and emergency management, through public
ontologies such as beAWARE \cite{beaware}; the cross-cutting plumbing of data
catalogues, provenance, sensing, and time, through DCAT \cite{dcat3}, PROV-O
\cite{provo}, SOSA and SSN \cite{janowicz2019sosa,haller2019ssn}, and the Time
Ontology \cite{owltime}; and the foundation layer of HQDM
\cite{west2011hqdm} and BFO \cite{arp2015bfo,isoiec21838}. The same five-stage
pipeline runs on every pair, which is what makes the matrix a single method
rather than a collection of bespoke mappings.

\subsection{Licensing and reproducibility}
\label{sec:method-licence}

Not every target can be redistributed. Some standards, in particular ISO~15926
and the IEC grid model standards, are copyrighted, so they are referenced for
alignment and never republished: the crosswalk records the correspondence to
their terms without carrying their text. Openly licensed sources are used
directly, and IES itself is available with its code under the MIT licence and
its documentation under the Open Government Licence \cite{ies4}, which makes a
published, attributed extension a legitimate contribution back to the standard
rather than an appropriation of it. The building and incident-management
extensions we use are open source under the Apache~2.0 licence, with their
documentation under the Open Government Licence
\cite{iesbuilding,lisaschema}. For every target the build records the version,
the namespace IRI, and the licence, and for the IES-side files we record the
source commit and check that our local copy is byte-identical to it; the full
inventory is given in \cref{sec:eval}. Every number reported in this paper is produced by the
pipeline from these recorded inputs, and the commands that generate each
artifact are stored alongside the artifact so that any figure can be
regenerated.

\section{Evaluation}
\label{sec:eval}

This section is structured around the questions the method is meant to answer,
not around the chronology of the runs. Every number is produced by the
reproducible build described in \cref{sec:method}, run locally on an Apple M3
Max with the engine built in release mode; the alignment steps need no network.
Each timing is the median of three runs, and each pair carries a command that
regenerates its artifacts. The matrix covers thirteen ontology pairs in seven
sector groups. Five further targets could not be obtained as openly licensed OWL
files in this offline run and are listed with their reasons in
\cref{sec:eval-skip}; the building anchor is carried by SAREF4BLDG, BOT, and
Brick rather than ifcOWL.

The evaluation addresses four questions:

\begin{description}
  \item[Q1 (Inventory).] What is each ontology, at which version, under which
    licence, and how large is it?
  \item[Q2 (Alignment).] How many correspondences does the engine propose per
    pair, is the result deterministic, and how fast is it?
  \item[Q3 (Certification).] How many proposed correspondences are held at their
    claimed strength, how many are downgraded, and how much of the source
    vocabulary is left unmapped?
  \item[Q4 (Coherence).] Does the merged ontology stay coherent and
    conservative, and how much of that is due to review and how much to
    certification?
\end{description}

\subsection{Q1: Ontology inventory}
\label{sec:eval-inventory}

The source side is IES in one of three profiles, and the profiles do not share
a release. The core profile, used for the energy, geospatial, catalogue,
plumbing, and foundation targets, is IES Common~5.0.3 (\num{511} classes,
\num{206} properties). The building profile is IES~4 together with an
open-source building extension, taken with the copy of the IES~4 schema it is
published alongside \cite{iesbuilding} (\num{706} classes, \num{249}
properties). The emergency-management profile is the same IES~4 schema
together with an open-source incident-management schema and hazards vocabulary
\cite{lisaschema} (\num{539} classes, \num{231} properties). That copy of
IES~4 is not identical to the 4.3.1 release \cite{ies4}: \num{26} of its
classes are absent from the release and \num{18} classes of the release are
absent from it. The building results therefore speak for this revision of
IES~4 and its building extension, not for IES~4.3.1 or IES~5. All three profiles are
RDF Schema vocabularies with no disjointness axioms; to run the OWL tools we add
an \texttt{owl:Class} typing to every named \texttt{rdfs:Class}, the only change
made to any input. The targets are listed in \cref{tab:inventory}, with their
disjointness axiom counts, which matter for what the coherence test can detect
(\cref{sec:eval-blind}). Copyrighted standards, namely the IEC grid model and,
in the skipped set, ISO~15926, are referenced for alignment only and never
redistributed; IES itself is openly licensed, which is what makes a published
extension a contribution back to the standard \cite{ies4}.

\begin{table}[htbp]
  \centering
  \caption{Target ontologies in the crosswalk matrix, with version, licence,
  size, and number of disjointness axioms. IES-side sizes are given per profile
  in the text; no IES profile has a disjointness axiom.}
  \label{tab:inventory}
  \small
  \begin{tabularx}{\linewidth}{l >{\raggedright\arraybackslash}X l r r r}
    \toprule
    Sector & Target (version) & Licence & Classes & Props & Disj. \\
    \midrule
    Built environment & SAREF4BLDG v1.1.2 / SAREF core v3.2.1 & ETSI (free to use) & 163 & 347 & 3 \\
    Built environment & BOT 0.3.2 & CC-BY & 8 & 17 & 18 \\
    Built environment & Brick 1.4.2 & BSD-3-Clause & 1794 & 153 & 30 \\
    Energy & IEC CGMES / CIM Core Equipment 3.0.0 & IEC (reference only) & 210 & 0 & 0 \\
    Energy & Open Energy Ontology 2.14.0 & CC0-1.0 & 1623 & 199 & 47 \\
    Infrastructure & OGC GeoSPARQL 1.1 & OGC document licence & 6 & 54 & 1 \\
    Emergency mgmt & beAWARE ontology v1.0 & Apache-2.0 & 63 & 83 & 3 \\
    Data catalogue & W3C DCAT v3 (2024) & W3C document licence & 9 & 39 & 0 \\
    Plumbing & SOSA / SSN (2017) & W3C document licence & 22 & 38 & 0 \\
    Plumbing & PROV-O (2013) & W3C document licence & 31 & 50 & 4 \\
    Plumbing & OWL-Time (2017) & W3C document licence & 20 & 58 & 1 \\
    Foundation & HQDM 1.0 (MagmaCore) & Apache-2.0 & 229 & 86 & 14 \\
    Foundation & BFO 2020 & CC-BY-4.0 & 36 & 40 & 10 \\
    \bottomrule
  \end{tabularx}
\end{table}

\subsection{Q2: Alignment}
\label{sec:eval-align}

\Cref{tab:align} reports, per pair, the number of engine candidates at the fixed
confidence threshold, how many were rejected on review, most of them same-word homonyms,
how many mappings the reviewer added that name matching could not see, and the
median alignment time. Every pair produced the same candidate set in each of its
three runs, so a determinism column would be uniformly positive and is omitted.
The candidate counts are small against large source vocabularies, which is
expected: most classes in one ontology have no counterpart in another, and three
pairs (GeoSPARQL, DCAT, OWL-Time) returned no automatic candidate at all, with
their correspondences found only where a reviewer looked for bridge points. The
homonym rejections catch a trap that label-based matching falls into, for
instance an IES temporal State against a device's on or off State, or a
building-component Element against the IES top-level Element.

\begin{table}[htbp]
  \centering
  \caption{Alignment summary per pair. Threshold, informative scoring, and
  one-to-one cardinality are fixed across all pairs (\cref{sec:method-align}).
  All pairs were deterministic across three runs. Of the seven
  rejections, five are same-word homonyms, one pairs differently named classes
  with different meanings (IES \texttt{SystemState} and Brick
  \texttt{System\_Status}), and one maps onto \texttt{owl:Thing}.}
  \label{tab:align}
  \small
  \begin{tabularx}{\linewidth}{X r r r r}
    \toprule
    Pair (IES to) & Candidates & Rejected & Added & Align med.\ (s) \\
    \midrule
    SAREF4BLDG  & 9  & 1 & 2 & 0.82 \\
    BOT         & 4  & 1 & 1 & 0.22 \\
    Brick       & 19 & 2 & 0 & 2.23 \\
    IEC CGMES   & 2  & 0 & 0 & 0.33 \\
    Open Energy Ontology & 22 & 0 & 0 & 1.55 \\
    GeoSPARQL   & 0  & 0 & 2 & 0.05 \\
    beAWARE     & 5  & 1 & 0 & 0.14 \\
    DCAT        & 0  & 0 & 0 & 0.07 \\
    SOSA / SSN  & 2  & 0 & 2 & 0.08 \\
    PROV-O      & 6  & 2 & 2 & 0.11 \\
    OWL-Time    & 0  & 0 & 2 & 0.06 \\
    HQDM        & 19 & 0 & 0 & 0.26 \\
    BFO         & 2  & 0 & 0 & 0.09 \\
    \bottomrule
  \end{tabularx}
\end{table}

\subsection{Q3: Certification outcomes}
\label{sec:eval-certify}

\Cref{tab:certify} gives, for each pair, how many correspondences certification
checked and how many it downgraded, first on the raw engine crosswalk and then
on the reviewed one, together with the source terms the reviewed crosswalk
leaves unmapped and the median certification time. Certification downgraded
\num{56} of the \num{90} raw correspondences and removed none: it changes the
strength of a correspondence, never its presence. Downgrading is the common case
on raw output because a label-driven matcher proposes \texttt{exactMatch} on a
shared lexical surface that the two hierarchies rarely make co-extensive. Brick
loses \num{16} of \num{19} raw equivalences, the Open Energy Ontology \num{13}
of \num{22}, HQDM \num{12} of \num{19}, and PROV-O \num{5} of \num{6}, to
broader, narrower, or merely related relations. Review does not make
certification redundant: on the reviewed crosswalks it still downgrades
\num{35} of \num{94} correspondences, including \num{8} of Brick's \num{17} and
\num{10} of the Open Energy Ontology's \num{22}, because a reviewer who accepts
two classes as the same thing cannot see that the hierarchies support only a
subsumption. The unmapped figures count the source terms in the materialised
hierarchy that have no row. They include terms that the building extension uses
as subclasses without declaring them as classes, which is why the building
pairs leave more terms unmapped than the profile declares classes. They are
non-overlap by design, not missed work. The data-catalogue pair is the limiting
case: it has no class-level overlap with IES at all and is omitted.

\begin{table}[htbp]
  \centering
  \caption{Certification outcomes per pair, on the raw engine crosswalk and on
  the reviewed crosswalk: correspondences checked and downgraded because the
  hierarchy did not support the claimed strength. A dash marks a pair with no
  engine candidates. Unmapped counts source terms of the materialised hierarchy
  with no row in the reviewed crosswalk; the time is the median of three
  certification runs on the reviewed crosswalk. Generated from the run
  artifacts.}
  \label{tab:certify}
  \small
  \begin{tabularx}{\linewidth}{X r r r r r r}
    \toprule
    & \multicolumn{2}{c}{Raw} & \multicolumn{2}{c}{Reviewed} & & \\
    \cmidrule(lr){2-3}\cmidrule(lr){4-5}
    Pair (IES to) & Checked & Down. & Checked & Down. & Unmapped src & Med.\ (s) \\
    \midrule
    SAREF4BLDG & 9 & 5 & 10 & 3 & 715 & 0.30 \\
    BOT & 4 & 3 & 4 & 0 & 721 & 0.35 \\
    Brick & 19 & 16 & 17 & 8 & 708 & 0.44 \\
    IEC CGMES & 2 & 0 & 2 & 0 & 509 & 0.27 \\
    Open Energy Ontology & 22 & 13 & 22 & 10 & 489 & 0.56 \\
    GeoSPARQL & -- & -- & 2 & 0 & 509 & 0.18 \\
    beAWARE & 5 & 1 & 4 & 0 & 536 & 0.21 \\
    SOSA / SSN & 2 & 0 & 4 & 0 & 507 & 0.18 \\
    PROV-O & 6 & 5 & 6 & 2 & 505 & 0.19 \\
    OWL-Time & -- & -- & 2 & 0 & 509 & 0.11 \\
    HQDM & 19 & 12 & 19 & 12 & 492 & 0.58 \\
    BFO & 2 & 1 & 2 & 0 & 509 & 0.11 \\
    \midrule
    Total & 90 & 56 & 94 & 35 & & \\
    \bottomrule
  \end{tabularx}
\end{table}

\subsection{Q4: Coherence and conservativity}
\label{sec:eval-coherence}

\begin{figure}[t]
  \centering
  \includegraphics[width=0.84\textwidth]{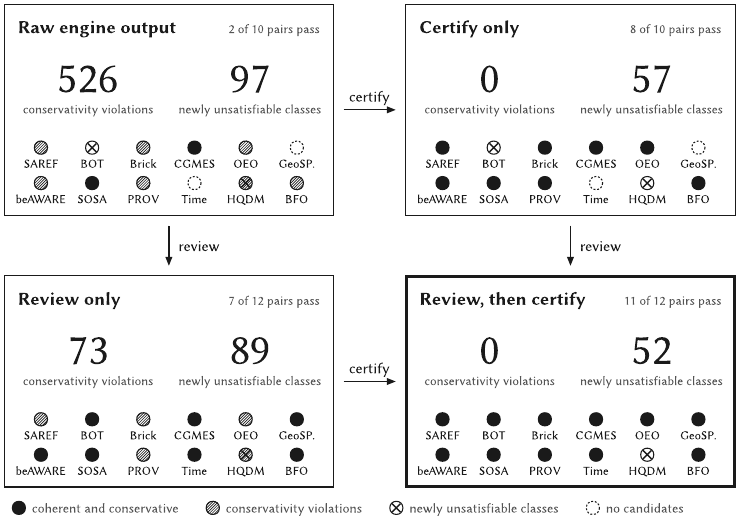}
  \caption{What review and certification each contribute, summed over the
  twelve pairs of \cref{tab:coherence}. Each panel is one crosswalk condition.
  The large numbers are the total conservativity violations and the total newly
  unsatisfiable classes, and each token is one pair. Certification alone clears
  every violation but keeps BOT's homonym; review alone removes the homonym but
  leaves violations on five pairs. Every newly unsatisfiable class in the lower
  panels comes from HQDM.}
  \label{fig:ablation}
\end{figure}

\Cref{tab:coherence} reports, for each pair and each of the four crosswalks of
\cref{sec:method-reason}, the number of classes that the merge makes
unsatisfiable although they are satisfiable in their own ontology, and the
number of conservativity violations forced onto the source and onto the target
vocabulary, and \cref{fig:ablation} sums the table over all pairs. Three
results stand out.

First, certification on its own removes every conservativity violation. The raw
engine crosswalks force \num{526} new subsumptions onto the two sides' own
vocabularies across seven pairs, as many as \num{187} on Brick's side alone, and
the same crosswalks after certification force none, on every pair. Review on its
own is not enough: the reviewed crosswalks still force \num{73} violations on
five pairs, from equivalences the reviewer kept that the two hierarchies support
only as subsumptions. This agreement is expected
rather than surprising, because certification is designed to keep only the
relations that the two hierarchies jointly support. What the reasoner adds is an
independent check of that design under the full OWL semantics of both inputs,
and it agrees on every pair. Certification also does this without deleting
anything: the certified crosswalks keep every reviewed correspondence and promote
to the same number of bridge axioms, except where a correspondence is weakened
to a related match, which promotes to none.

Second, certification on its own does not remove unsatisfiability, because it
cannot see disjointness. The BOT pair shows the case cleanly. The engine proposes
that the IES top-level \texttt{Element} is the same as BOT's \texttt{Element}, a
building component, which BOT declares disjoint from \texttt{bot:Zone}, the
superclass of \texttt{bot:Building}. The engine also equates the IES building,
storey, and space classes, which sit under the IES \texttt{Element}, with their
BOT namesakes, which sit under \texttt{bot:Zone}. Each of those classes is then
forced under both sides of the disjointness, and the raw merge has \num{8} new
unsatisfiable classes: three in BOT and five in IES, the IES space's two
subclasses among them. Certification compares only subsumption, and nothing in
either hierarchy refutes the \texttt{Element} equivalence, so it keeps it as an exact match. It
does weaken the three building-part equivalences to subsumptions, which frees
the BOT classes, but the five IES classes still sit under both
\texttt{bot:Element} and \texttt{bot:Zone} and stay unsatisfiable. Only the
reviewer's rejection of the homonym, which maps \texttt{bot:Element} to the IES
built-entity class instead, takes the count to zero. Review and
certification therefore catch different failures, and the reviewed and certified
crosswalk is the only one of the four conditions that is both coherent and
conservative on BOT, SAREF4BLDG, and Brick at once.

Third, the HQDM pair is not certified under any of the four conditions. HQDM is
already incoherent on its own: \num{39} of its \num{229} classes are
unsatisfiable before any alignment is added, as our earlier work reported
\cite{rovai2026coherence}. The raw crosswalk makes a further \num{89} classes
unsatisfiable, \num{73} in IES and \num{16} in HQDM, and the certified crosswalk
still makes \num{52} IES classes unsatisfiable. The cause is a single
correspondence. Certification holds \texttt{ies:Asset} as an exact match for
\texttt{hqdm:asset}, and \texttt{hqdm:asset} is one of the \num{39} classes that
HQDM cannot populate, so the equivalence empties \texttt{ies:Asset} and every IES
class beneath it. Re-reasoning the certified crosswalk without that one row gives
no new unsatisfiable class. We report this as a diagnosis, not as a repaired
result, because the row was removed after the fact on one pair. It points to a
rule the certification stage lacks: a correspondence onto a class that is
unsatisfiable in its own ontology says nothing true and should be refused before
any strength is assigned. Our earlier work flagged exactly this kind of mapping
onto \texttt{hqdm:asset} as vacuously true when produced by other matchers, and
its orientation search, which uses the reasoner as an oracle during the search
rather than once at the end, produced a certified 21-axiom IES-to-HQDM bridge
\cite{rovai2026coherence}. For HQDM, that bridge remains the certified crosswalk;
the lighter pipeline of this paper does not reproduce it.

\begin{table}[htbp]
  \centering
  \caption{Coherence and conservativity under four crosswalks per pair. U is the
  number of classes made unsatisfiable by the merge that are satisfiable in their
  own ontology; C is the number of conservativity violations on the source and
  target vocabularies (source/target). A dash marks a pair with no engine
  candidates. On an incoherent merge, C counts only classes that stay
  satisfiable. DCAT has no class overlap and is omitted. HermiT time is the
  median of three runs on the reviewed merge.}
  \label{tab:coherence}
  \footnotesize
  \setlength{\tabcolsep}{4pt}
  \begin{tabular}{l *{4}{>{\centering\arraybackslash}p{0.55cm} >{\centering\arraybackslash}p{1.2cm}} r}
    \toprule
    & \multicolumn{2}{c}{Raw} & \multicolumn{2}{c}{Certified raw}
    & \multicolumn{2}{c}{Reviewed} & \multicolumn{2}{c}{Reviewed + cert.} & \\
    \cmidrule(lr){2-3}\cmidrule(lr){4-5}\cmidrule(lr){6-7}\cmidrule(lr){8-9}
    Pair (IES to) & U & C & U & C & U & C & U & C & HermiT (s) \\
    \midrule
    SAREF4BLDG  & 0  & 54/77  & 0  & 0/0 & 0  & 17/0  & 0  & 0/0 & 2.13 \\
    BOT         & 8  & 0/0    & 5  & 0/0 & 0  & 0/0   & 0  & 0/0 & 1.14 \\
    Brick       & 0  & 25/187 & 0  & 0/0 & 0  & 18/10 & 0  & 0/0 & 1.24 \\
    IEC CGMES   & 0  & 0/0    & 0  & 0/0 & 0  & 0/0   & 0  & 0/0 & 0.80 \\
    Open Energy Ont. & 0 & 23/1 & 0 & 0/0 & 0 & 0/1   & 0  & 0/0 & 68.26 \\
    GeoSPARQL   & -- & --     & -- & --  & 0  & 0/0   & 0  & 0/0 & 1.01 \\
    beAWARE     & 0  & 28/0   & 0  & 0/0 & 0  & 0/0   & 0  & 0/0 & 0.76 \\
    SOSA / SSN  & 0  & 0/0    & 0  & 0/0 & 0  & 0/0   & 0  & 0/0 & 1.07 \\
    PROV-O      & 0  & 0/3    & 0  & 0/0 & 0  & 0/2   & 0  & 0/0 & 1.07 \\
    OWL-Time    & -- & --     & -- & --  & 0  & 0/0   & 0  & 0/0 & 0.63 \\
    HQDM        & 89 & 25/0   & 52 & 0/0 & 89 & 25/0  & 52 & 0/0 & 0.59 \\
    BFO         & 0  & 103/0  & 0  & 0/0 & 0  & 0/0   & 0  & 0/0 & 0.39 \\
    \bottomrule
  \end{tabular}
\end{table}

With review and certification together, eleven of the twelve pairs with class
overlap are coherent and conservative, including all three building pairs; HQDM
is the exception. The BFO pair is coherent and conservative but thin: BFO uses
opaque numeric identifiers, so automatic name matching finds little, and the
reviewed crosswalk holds one broad and one close match rather than any
equivalence. A deeper IES-to-BFO mapping is manual, community work, and the
method reaches only its surface here. Reasoning is fast on every pair except the
Open Energy Ontology, whose merge takes about a minute under HermiT.

\subsection{Q5: From certified pairs to a certified hub}
\label{sec:eval-hub}

An ecosystem of connected twins is not a set of separate pairs. Every crosswalk
into one common ontology is loaded alongside every other, and a consumer of one
sector's data reasons over all of them at once. We therefore merged each IES
profile with all of its targets and all of their reviewed and certified
crosswalks into a single ontology, and ran \cref{alg:accept} on the whole hub:
newly unsatisfiable classes against each input on its own, and conservativity
violations against each input's own vocabulary. We also counted what the hub is
for: subsumptions between classes of two different targets, neither of them an
IES class, that the hub entails and that nobody wrote down.
\Cref{tab:hub} gives the results.

\begin{table}[htbp]
  \centering
  \caption{Reasoning over whole hubs. Inputs counts the IES profile and its
  targets; rows are the reviewed and certified crosswalk rows merged in. Unsat
  counts classes unsatisfiable in the hub but satisfiable in their own input;
  violations are conservativity violations summed over all inputs; cross-links
  are entailed subsumptions between classes of two different targets, neither
  an IES class. HQDM, already incoherent on its own, is left out of the core
  hub. Generated from the run artifacts.}
  \label{tab:hub}
  \small
  \begin{tabularx}{\linewidth}{X r r r r r r}
    \toprule
    Hub & Inputs & Rows & Unsat & Violations & Cross-links & HermiT (s) \\
    \midrule
    Building, as certified pair by pair & 4 & 31 & 5 & 0 & 280 & 1.7 \\
    Building, \texttt{Floor} row rejected & 4 & 30 & 0 & 0 & 290 & 1.4 \\
    Core, all seven targets & 8 & 40 & \multicolumn{3}{c}{inconsistent} & 1.3 \\
    Core without BFO & 7 & 38 & 0 & 0 & 95 & 451.4 \\
    Core without the Open Energy Ontology & 7 & 18 & 0 & 0 & 0 & 12.8 \\
    \bottomrule
  \end{tabularx}
\end{table}

Pairwise certification does not compose. Each of the three building pairs is
coherent and conservative on its own (\cref{tab:coherence}), yet the building
hub makes five Brick classes unsatisfiable: \texttt{Floor}, \texttt{Storey},
\texttt{Basement}, \texttt{Parking\_Level}, and \texttt{Rooftop}. The cause is
a homonym in the Brick crosswalk that only a third ontology can expose. The
reviewer kept the engine's match between the IES building extension's
\texttt{Floor} and Brick's \texttt{Floor}, with the note that both ``define a
storey-level floor surface of a building'', and certification weakened it to a
subsumption. But the two are different things. The IES \texttt{Floor} is a
structural component, a built entity that occupants walk on; Brick's
\texttt{Floor} is a level of a building, declared equivalent to
\texttt{brick:Storey}, with \texttt{Basement}, \texttt{Parking\_Level}, and
\texttt{Rooftop} beneath it. IES has no disjointness axioms, so the IES--Brick
merge is coherent and the error is invisible. The BOT crosswalk maps the IES
built entity to \texttt{bot:Element} and the IES storey to \texttt{bot:Storey},
and BOT declares elements and zones disjoint. In the hub, Brick's levels are
forced under both, and become empty. Rejecting that one row makes the building
hub coherent and conservative on all four inputs.

The repaired building hub entails \num{290} cross-links between targets, none
of them stated by anyone. Brick's room types, from offices to kitchens to
storage rooms, fall under SAREF4BLDG's building space and under BOT's space and
zone; Brick's storeys fall under BOT's storey. In a random sample of eight
links from each repaired hub none was wrong, although many of the core hub's
links land on PROV-O's very general entity class and say little. These are entailments of the
certified crosswalks, not reviewed correspondences, and a hub that publishes
them should mark them as such, but they are what an ecosystem of connected
twins is meant to deliver: a building ontology connected to IES is thereby
connected to every other building ontology connected to IES.

The core hub fails for a different reason, one that no crosswalk can fix. With
all seven targets it is inconsistent, and it stays inconsistent with every
crosswalk row removed. The cause is two published standards: the Open Energy
Ontology together with BFO~2020 is inconsistent on its own, and with the Open
Energy Ontology's \num{232} named individuals left out the pair still makes
three classes unsatisfiable. The Open Energy Ontology carries its own copy of
\num{35} BFO classes, which suggests a version mismatch, though we have not
isolated the axioms responsible. Either target can join the hub, but not both.
Without BFO, the core hub is coherent and conservative on all seven inputs and
entails \num{95} cross-links, mostly energy-system classes placed under PROV-O's
entity, agent, organisation, person, and location; without the Open Energy
Ontology it is coherent and conservative and entails none. The core hub is also
the expensive one: classifying it with the Open Energy Ontology in place took
HermiT over seven minutes, against seconds for every other merge.

\subsection{What the acceptance test cannot see}
\label{sec:eval-blind}

A coherence test is only as sharp as the axioms it can contradict. No IES
profile carries a disjointness axiom, so every unsatisfiable class in our merges
comes from a disjointness or an unsatisfiable class on the target side. Three
targets, the IEC grid model, DCAT, and SOSA/SSN, carry no disjointness axiom
either (\cref{tab:inventory}), so for those pairs the coherence half of the test
has almost nothing to fire on, and a zero there is weak evidence; only the
conservativity half has teeth. More generally, a crosswalk can be coherent and
conservative and still wrong. The certified HQDM crosswalk holds
\texttt{ies:Event} as an exact match for \texttt{hqdm:event}. Our earlier work
argues that this is a false friend: an IES event is a durative happening with
participants, an HQDM event is an instantaneous boundary, and the true
counterpart is \texttt{hqdm:activity} \cite{rovai2026coherence}. Neither the
hierarchy test nor the reasoner objects, because nothing in either ontology's
axioms forbids the identification. Certification bounds the logical damage a
crosswalk can do; it does not establish that the crosswalk means the right thing,
which remains the reviewer's job.

\subsection{Building anchor: the extension step}
\label{sec:eval-anchor}

The anchor case measures the specific claim that extending an IES building
vocabulary for regional retrofit is a quick and checkable step rather than a
costly manual step. We built a small attributed OWL module that reuses the
open IES building classes as parents and ranges and adds a retrofit-planning
layer: a project and a regional programme, a measure hierarchy with fabric,
heating, and renewable-generation branches, and a recommendation type that makes
the building-performance recommendation analysable at scale.
\Cref{tab:anchor} records the result. The engine's planner checks an extension
against the base by deriving, under a Horn rule approximation of OWL, every
consequence the extension adds over the base vocabulary. That check is sound in
one direction only: a new consequence is a real change of meaning, while finding
none does not prove the extension conservative in OWL. We therefore also ran
HermiT on the merge as an independent check.

\begin{table}[htbp]
  \centering
  \caption{Building-ontology extension for regional retrofit, measured on an
  Apple M3 Max. Plan, apply, diff, and reasoning times are medians of three runs.}
  \label{tab:anchor}
  \small
  \begin{tabularx}{\linewidth}{X r}
    \toprule
    Measure & Value \\
    \midrule
    Classes added                               & 14 \\
    Object properties added                     & 7 \\
    Datatype properties added                   & 10 \\
    Triples added / base triples dropped        & 123 / 0 \\
    New consequences over the base vocabulary (rule approximation) & none found \\
    Plan risk / new TBox defects / ABox         & low / 0 / consistent (90 individuals) \\
    Plan time, median (s)                       & 0.72 \\
    Apply time, median (s)                      & 0.046 \\
    Diff time, median (s)                       & 0.039 \\
    Independent reasoner (HermiT) new unsat     & 0 \\
    HermiT merge-check time, median (s)         & 0.33 \\
    \bottomrule
  \end{tabularx}
\end{table}

The module ships with a SHACL data contract for building-performance certificate
data: a certificate must carry one rating band in the A to G range, a score on
the 1 to 100 scale, a positive floor area, and a non-negative emissions figure.
Validated against a mixed sample under the engine's SHACL evaluator, the contract
returned non-conformance with five violations, and all five fell on exactly the
two records built to fail, three on a malformed certificate and two on a
malformed recommendation. A genuine open certificate (a mid-terrace house, band
D with potential B, \num{135} square metres, \SI{7.0}{\tonne} of
CO\textsubscript{2} a year) and every valid synthetic record passed. The
contract accepts the good records and rejects the bad ones, and it names the
constraint each bad record breaks. The extension is version 0.1.0, deliberately
small, and models the retrofit-planning layer rather than the assessment process
the base ontology already covers.

\subsection{Targets not obtained in this run}
\label{sec:eval-skip}

Five targets could not be fetched as openly licensed OWL in this offline run and
are reported rather than silently dropped. ifcOWL requires buildingSMART
distribution channels, and the built-environment anchor is carried by
SAREF4BLDG, BOT, and Brick instead. HealthDCAT-AP was not obtainable as an OWL
or SHACL artifact, and catalogue plumbing is carried by DCAT. ISO~15926 is
copyrighted with no openly licensed OWL available, and the 4D commitments it
shares are reached through HQDM. BORO and IDEAS likewise had no openly fetchable
reference OWL, and the BORO method is reached through HQDM, the 4D upper
ontology built with it.

\section{Discussion}
\label{sec:discussion}

The method rests on one design choice: proposing a correspondence and accepting
it are separate steps, and acceptance is a measured property of the merged
ontology rather than a confidence score or a vote. The criteria themselves,
coherence and conservativity, come from earlier work on logic-based alignment
repair \cite{meilicke2009revision,jimenezruiz2011logmap,solimando2014conservativity}.
What this paper adds is their application to IES across seven sector groups, a
measurement of which stage of a practical pipeline does the work, and the step
from pairs to hubs.

The hub result is the one that matters most for an ecosystem of connected
twins. Certifying each crosswalk on its own is necessary and not enough: the
building hub failed although every building pair passed, because the axiom that
exposed the Brick homonym lived in BOT, a third ontology that neither the IES
nor the Brick side contains. In a federation, every new crosswalk is checked
against every ontology already connected, whether or not anyone intends it, so
the acceptance test has to run on the hub each time a crosswalk joins. The core
hub shows the other half of the same lesson. Two standards that are each
coherent, and each certified against IES, cannot be loaded together, and no
crosswalk repair can change that; a hub test is also a test of the standards it
connects. Both checks took seconds, except where the Open Energy Ontology was
involved, so running them on every change is affordable.

That measurement is the clearest result. Review and certification fail in
different ways, and each covers the other's blind spot. Certification reads the
two class hierarchies and sets each correspondence to the strength they jointly
support; on its own it removed all \num{526} conservativity violations in the raw
crosswalks, on every pair, without deleting a correspondence. It cannot see
disjointness, so it left the BOT homonym in place and five classes
unsatisfiable. Review catches homonyms, because a reviewer reads the definitions,
but a reviewer who accepts two classes as the same thing has no way to see that
the two hierarchies support only a subsumption, and the reviewed crosswalks still
carried \num{73} violations. Neither stage alone made the three building pairs
coherent and conservative; both together did (\cref{tab:coherence}).

The HQDM pair marks the limit of the pipeline. A single held equivalence onto a
class that HQDM cannot populate made \num{52} IES classes unsatisfiable, and
neither review nor certification caught it. Two lessons follow. The certification
stage should refuse any correspondence whose target is unsatisfiable in its own
ontology, a cheap check that needs one reasoner call per input. And when the
ontologies being joined are themselves incoherent, a final check is not enough;
the orientation search of our earlier work, which consults the reasoner during
the search, is the tool that produced a certified IES-to-HQDM bridge
\cite{rovai2026coherence}.

Several findings are negative and are reported as such. Energy is a gap, not a
failure: neither the grid model nor the energy ontology shares IES's core
vocabulary, and the only certified matches are generic entities such as
currency, organisations, and units, so energy interoperability needs an IES
energy extension rather than a better matcher (\cref{tab:certify}). The data
catalogue has no class-level overlap with IES and sits beside it as plumbing.
Alignment to BFO is thin because BFO uses opaque numeric identifiers, so only
broad and close matches were found.

The claims are bounded in four further ways. The matcher proposes few
correspondences against large vocabularies, so the method certifies the
correspondences it finds and promises nothing about coverage. Where neither
ontology has disjointness axioms, as for three of our targets, the coherence test
can barely fail, and only the conservativity test carries weight
(\cref{sec:eval-blind}). Coherent and conservative is not the same as correct:
the certified HQDM crosswalk still equates an IES event with an HQDM event, a
false friend that no axiom in either ontology forbids. The building pairs
use IES~4 with its building extension, so they say nothing yet about IES~5.
Finally, IES's own documentation says the ontology ``is not intended for
reasoning / inference'' \cite{ies4}. We reason over it not to draw new facts
about the world but to find out what a merge already commits a consumer to,
which is what any system that loads a crosswalk does whether or not the
standard intends it.

The broader aim is a single IES hub whose crosswalks carry their acceptance
evidence with them. The same pipeline that reaches a building vocabulary reaches
temporal structure through the Time Ontology \cite{owltime}, provenance-carrying
data contracts through PROV-O \cite{provo}, sensor and Earth-observation data
through SOSA and SSN \cite{janowicz2019sosa,haller2019ssn}, geospatial asset data
through GeoSPARQL \cite{battle2012geosparql}, and the foundation layer of HQDM and
BFO \cite{west2011hqdm,arp2015bfo}. Each of these is a row in the matrix of
\cref{sec:method-matrix}, measured with the same four conditions, rather than a
separate research programme.

\section{Conclusion}
\label{sec:conclusion}

We applied the coherence and conservativity acceptance test to crosswalks from
the Information Exchange Standard into thirteen ontologies across seven sector
groups, with the building-ontology landscape as the anchor. We cited that
landscape as the prior art it is and scoped our novelty to the IES side: we are
not aware of prior published certified alignments between IES and the
established building-ontology landscape. Measuring each pair under four
crosswalks showed that certification and human review catch different failures,
certification removing every conservativity violation and review removing the
homonyms that cause unsatisfiability, and that together they make eleven of the
twelve overlapping pairs, including all three building pairs, coherent and
conservative. The twelfth, HQDM, stays incoherent because of one correspondence
onto a class HQDM cannot itself populate, which motivates a refusal rule the
pipeline lacks. A retrofit extension of the IES building vocabulary, \num{14}
classes and \num{17} properties, was planned in under a second, applied in under
a tenth of a second, and added no unsatisfiable class under an independent
reasoner. Follow-up work should add the refusal rule and run orientation search on
the pairs with incoherent inputs, repeat the building pairs on IES~5, extend the
anchor to bulk open data, deepen the IES-to-BFO mapping by hand, and scope an IES
energy extension.

\bibliographystyle{elsarticle-num-names}
\bibliography{refs}

@misc{rovai2026coherence,
  title        = {Consistency Is Not Coherence: Orientation Search for Certified Alignments Between 4D Defence Upper Ontologies},
  author       = {Rovai, Fabio},
  year         = {2026},
  eprint       = {2608.21914},
  archivePrefix= {arXiv},
  primaryClass = {cs.AI},
  note         = {Long paper accepted at the Ontology Matching workshop (OM 2026), co-located with ISWC 2026},
  url          = {https://arxiv.org/abs/2608.21914}
}

@misc{dilworth2026defence,
  title        = {Improving Interoperability among Defence and National Security Ontologies: Analysis and Evaluation Tasks},
  author       = {Dilworth, Jonathon and Cotovio, Pedro Giesteira and Herron, David and Cripps, Paul and Dewdney, Nigel and Pesquita, Catia and Jim{\'e}nez-Ruiz, Ernesto},
  year         = {2026},
  eprint       = {2608.05867},
  archivePrefix= {arXiv},
  primaryClass = {cs.AI},
  note         = {Accepted at the Resource Track of the 25th International Semantic Web Conference (ISWC 2026), Bari, Italy},
  url          = {https://arxiv.org/abs/2608.05867}
}

@misc{rovai2026openontologies,
  title        = {Open Ontologies: Tool-Augmented Ontology Engineering with Stable Matching Alignment},
  author       = {Rovai, Fabio},
  year         = {2026},
  eprint       = {2605.09184},
  archivePrefix= {arXiv},
  primaryClass = {cs.AI},
  note         = {Software released under the MIT licence},
  url          = {https://arxiv.org/abs/2605.09184}
}

@inproceedings{jimenezruiz2011logmap,
  title        = {{LogMap}: Logic-Based and Scalable Ontology Matching},
  author       = {Jim{\'e}nez-Ruiz, Ernesto and Cuenca Grau, Bernardo},
  booktitle    = {The Semantic Web -- ISWC 2011},
  series       = {Lecture Notes in Computer Science},
  pages        = {273--288},
  publisher    = {Springer},
  year         = {2011},
  doi          = {10.1007/978-3-642-25073-6_18}
}

@article{meilicke2009revision,
  title        = {Reasoning Support for Mapping Revision},
  author       = {Meilicke, Christian and Stuckenschmidt, Heiner and Tamilin, Andrei},
  journal      = {Journal of Logic and Computation},
  volume       = {19},
  number       = {5},
  pages        = {807--829},
  year         = {2009},
  doi          = {10.1093/logcom/exn047}
}

@inproceedings{solimando2014conservativity,
  title        = {Detecting and Correcting Conservativity Principle Violations in Ontology-to-Ontology Mappings},
  author       = {Solimando, Alessandro and Jim{\'e}nez-Ruiz, Ernesto and Guerrini, Giovanna},
  booktitle    = {The Semantic Web -- ISWC 2014},
  series       = {Lecture Notes in Computer Science},
  pages        = {1--16},
  publisher    = {Springer},
  year         = {2014},
  doi          = {10.1007/978-3-319-11915-1_1}
}

@article{solimando2017minimizing,
  title        = {Minimizing Conservativity Violations in Ontology Alignments: Algorithms and Evaluation},
  author       = {Solimando, Alessandro and Jim{\'e}nez-Ruiz, Ernesto and Guerrini, Giovanna},
  journal      = {Knowledge and Information Systems},
  volume       = {51},
  number       = {3},
  pages        = {775--819},
  year         = {2017},
  doi          = {10.1007/s10115-016-0983-3}
}

@article{rasmussen2021bot,
  title   = {{BOT}: The Building Topology Ontology of the {W3C} Linked Building Data Group},
  author  = {Rasmussen, Mads Holten and Lefran{\c{c}}ois, Maxime and Schneider, Georg Ferdinand and Pauwels, Pieter},
  journal = {Semantic Web},
  volume  = {12},
  number  = {1},
  pages   = {143--161},
  year    = {2021},
  doi     = {10.3233/SW-200385},
  url     = {https://doi.org/10.3233/SW-200385}
}

@article{balaji2018brick,
  title   = {{Brick}: Metadata schema for portable smart building applications},
  author  = {Balaji, Bharathan and Bhattacharya, Arka and Fierro, Gabriel and Gao, Jingkun and Gluck, Joshua and Hong, Dezhi and Johansen, Aslak and Koh, Jason and Ploennigs, Joern and Agarwal, Yuvraj and Berg{\'e}s, Mario and Culler, David and Gupta, Rajesh K. and Kj{\ae}rgaard, Mikkel Baun and Srivastava, Mani and Whitehouse, Kamin},
  journal = {Applied Energy},
  volume  = {226},
  pages   = {1273--1292},
  year    = {2018},
  doi     = {10.1016/j.apenergy.2018.02.091},
  url     = {https://doi.org/10.1016/j.apenergy.2018.02.091}
}

@article{pauwels2016ifcowl,
  title   = {{EXPRESS} to {OWL} for construction industry: Towards a recommendable and usable {ifcOWL} ontology},
  author  = {Pauwels, Pieter and Terkaj, Walter},
  journal = {Automation in Construction},
  volume  = {63},
  pages   = {100--133},
  year    = {2016},
  doi     = {10.1016/j.autcon.2015.12.003},
  url     = {https://doi.org/10.1016/j.autcon.2015.12.003}
}

@misc{saref,
  title        = {{SAREF}: The Smart Applications REFerence Ontology (core v4.1.1)},
  author       = {{ETSI}},
  howpublished = {ETSI SmartM2M standard, originated by Daniele, den Hartog and Roes (FOMI 2015)},
  year         = {2024},
  url          = {https://saref.etsi.org/core/}
}

@misc{miles2009skos,
  title        = {{SKOS} Simple Knowledge Organization System Reference},
  author       = {Miles, Alistair and Bechhofer, Sean},
  howpublished = {W3C Recommendation},
  year         = {2009},
  url          = {https://www.w3.org/TR/skos-reference/}
}

@article{matentzoglu2022sssom,
  title   = {A Simple Standard for Sharing Ontological Mappings ({SSSOM})},
  author  = {Matentzoglu, Nicolas and Balhoff, James P. and Bello, Susan M. and Bizon, Chris and Brush, Matthew and Callahan, Tiffany J. and others},
  journal = {Database},
  volume  = {2022},
  pages   = {baac035},
  year    = {2022},
  doi     = {10.1093/database/baac035},
  url     = {https://doi.org/10.1093/database/baac035}
}

@article{glimm2014hermit,
  title   = {{HermiT}: An {OWL} 2 Reasoner},
  author  = {Glimm, Birte and Horrocks, Ian and Motik, Boris and Stoilos, Giorgos and Wang, Zhe},
  journal = {Journal of Automated Reasoning},
  volume  = {53},
  number  = {3},
  pages   = {245--269},
  year    = {2014},
  doi     = {10.1007/s10817-014-9305-1},
  url     = {https://doi.org/10.1007/s10817-014-9305-1}
}

@book{euzenat2013ontologymatching,
  title     = {Ontology Matching},
  author    = {Euzenat, J{\'e}r{\^o}me and Shvaiko, Pavel},
  edition   = {2},
  publisher = {Springer},
  year      = {2013},
  doi       = {10.1007/978-3-642-38721-0},
  url       = {https://doi.org/10.1007/978-3-642-38721-0}
}

@article{janowicz2019sosa,
  title   = {{SOSA}: A lightweight ontology for sensors, observations, samples, and actuators},
  author  = {Janowicz, Krzysztof and Haller, Armin and Cox, Simon J.D. and Le Phuoc, Danh and Lefran{\c{c}}ois, Maxime},
  journal = {Journal of Web Semantics},
  volume  = {56},
  pages   = {1--10},
  year    = {2019},
  doi     = {10.1016/j.websem.2018.06.003},
  url     = {https://doi.org/10.1016/j.websem.2018.06.003}
}

@article{haller2019ssn,
  title   = {The modular {SSN} ontology: A joint {W3C} and {OGC} standard specifying the semantics of sensors, observations, sampling, and actuation},
  author  = {Haller, Armin and Janowicz, Krzysztof and Cox, Simon J.D. and Lefran{\c{c}}ois, Maxime and Taylor, Kerry and Le Phuoc, Danh and Lieberman, Joshua and Garc{\'i}a-Castro, Ra{\'u}l and Atkinson, Rob and Stadler, Claus},
  journal = {Semantic Web},
  volume  = {10},
  number  = {1},
  pages   = {9--32},
  year    = {2019},
  doi     = {10.3233/SW-180320},
  url     = {https://doi.org/10.3233/SW-180320}
}

@article{battle2012geosparql,
  title   = {Enabling the geospatial Semantic Web with Parliament and {GeoSPARQL}},
  author  = {Battle, Robert and Kolas, Dave},
  journal = {Semantic Web},
  volume  = {3},
  number  = {4},
  pages   = {355--370},
  year    = {2012},
  doi     = {10.3233/SW-2012-0065},
  url     = {https://doi.org/10.3233/SW-2012-0065}
}

@misc{provo,
  title        = {{PROV-O}: The {PROV} Ontology},
  author       = {Lebo, Timothy and Sahoo, Satya and McGuinness, Deborah},
  howpublished = {W3C Recommendation},
  year         = {2013},
  url          = {https://www.w3.org/TR/prov-o/}
}

@misc{owltime,
  title        = {Time Ontology in {OWL}},
  author       = {Cox, Simon and Little, Chris},
  howpublished = {W3C Recommendation, 19 October 2017},
  year         = {2017},
  url          = {https://www.w3.org/TR/2017/REC-owl-time-20171019/}
}

@misc{dcat3,
  title        = {Data Catalog Vocabulary ({DCAT}) - Version 3},
  author       = {Albertoni, Riccardo and Browning, David and Cox, Simon J. D. and Gonzalez Beltran, Alejandra and Perego, Andrea and Winstanley, Peter},
  howpublished = {W3C Recommendation, 22 August 2024},
  year         = {2024},
  url          = {https://www.w3.org/TR/vocab-dcat-3/}
}

@misc{beaware,
  title        = {Ontology-Based Representation of Crisis Management Procedures for Climate Events},
  author       = {Kontopoulos, Efstratios and Mitzias, Panagiotis and Mo{\ss}graber, J{\"u}rgen and Hertweck, Philipp and van der Schaaf, Hylke and Hilbring, D{\'e}sir{\'e}e and others},
  publisher    = {Zenodo},
  year         = {2018},
  doi          = {10.5281/zenodo.1243535},
  note         = {Recommended citation for The beAWARE Crisis Management Ontology, \url{https://github.com/beAWARE-project/ontology}}
}

@misc{ies4,
  title        = {{IES}: The Information Exchange Standard (IES4)},
  author       = {{Dstl and UK Government}},
  howpublished = {4D RDF Schema ontology; code MIT (v4.3.1), documentation under the Open Government Licence},
  year         = {2025},
  url          = {https://github.com/dstl/IES4}
}

@book{west2011hqdm,
  title     = {Developing High Quality Data Models},
  author    = {West, Matthew},
  publisher = {Morgan Kaufmann},
  year      = {2011},
  isbn      = {978-0-12-375106-5},
  url        = {https://www.oreilly.com/library/view/developing-high-quality/9780123751065/},
  doi          = {10.1016/C2009-0-30508-5}
}

@book{arp2015bfo,
  title     = {Building Ontologies with Basic Formal Ontology},
  author    = {Arp, Robert and Smith, Barry and Spear, Andrew D.},
  publisher = {MIT Press},
  year      = {2015},
  isbn      = {978-0-262-52781-1},
  url        = {https://mitpress.mit.edu/9780262527811/},
  doi          = {10.7551/mitpress/9780262527811.001.0001}
}

@misc{isoiec21838,
  title        = {{ISO/IEC} 21838-2:2021 Information technology --- Top-level ontologies (TLO) --- Part 2: Basic Formal Ontology (BFO)},
  author       = {{ISO/IEC}},
  year         = {2021},
  url          = {https://www.iso.org/standard/74572.html}
}

@misc{iesbuilding,
  title        = {{IES}~4 Building Extension (\texttt{ies-building1}), with the {IES}~4 Schema It Extends},
  author       = {{National Digital Twin Programme}},
  howpublished = {Open-source ontology files in the IRIS repository; code Apache-2.0, documentation Open Government Licence v3.0. Both files at commit 2db4240, the first released under these licences},
  year         = {2026},
  url          = {https://github.com/National-Digital-Twin/IRIS/blob/2db424030999f9db839920484a9c435bd07ab3eb/data-tools/data-pipeline/materialised-view-creation/src/create-view/ies-building.ttl}
}

@misc{lisaschema,
  title        = {Incident-Management Schema and Hazards Vocabulary for {IES}~4},
  author       = {{National Digital Twin Programme}},
  howpublished = {Open-source ontology files in the LISA repository (\texttt{schema/schema.ttl} at commit a38d2f6, \texttt{schema/Hazards/hazards.ttl} at commit d77f540); code Apache-2.0, documentation Open Government Licence v3.0},
  year         = {2025},
  url          = {https://github.com/National-Digital-Twin/LISA/tree/d77f540d146b3ede327a48fc4dfbd32dae34af51/schema}
}

@techreport{gemini2018,
  title       = {The {Gemini} Principles: Guiding Values for the National Digital Twin and Information Management Framework},
  author      = {Bolton, Alexandra and Butler, Lorraine and Dabson, Ian and Enzer, Mark and others},
  institution = {Centre for Digital Built Britain and Digital Framework Task Group},
  address     = {Cambridge},
  year        = {2018},
  doi         = {10.17863/CAM.32260},
  url         = {https://doi.org/10.17863/CAM.32260}
}

@techreport{hetherington2020imf,
  title       = {The Pathway Towards an Information Management Framework: A `Commons' for Digital Built Britain},
  author      = {Hetherington, James and West, Matthew},
  institution = {Centre for Digital Built Britain},
  address     = {Cambridge},
  year        = {2020},
  doi         = {10.17863/CAM.52659},
  url         = {https://doi.org/10.17863/CAM.52659}
}

@inproceedings{bailey2025iesbfo,
  title     = {Comparing Information Exchange Standard and Basic Formal Ontology Design Patterns},
  author    = {Bailey, Ian and Beverley, John and Blackmore, Helena and Cola, Andreas and Cripps, Paul and De Colle, Giacomo and Donato, Federico and Hicks, Amanda and Limbaugh, David and Milivinti, Elena and Partridge, Chris and Rafferty, Rebecca and Smith, Barry},
  booktitle = {Proceedings of the Joint Ontology Workshops (JOWO) - Episode XI, co-located with FOIS 2025},
  series    = {CEUR Workshop Proceedings},
  volume    = {4176},
  year      = {2025},
  address   = {Catania, Italy},
  url       = {https://ceur-ws.org/Vol-4176/foust-6.pdf}
}

@techreport{etsi2020saref4bldg,
  author      = {{ETSI}},
  title       = {{SmartM2M}; Extension to {SAREF}; Part 3: Building Domain},
  institution = {ETSI},
  type        = {Technical Specification},
  number      = {TS 103 410-3 V1.1.2},
  year        = {2020},
  url         = {https://www.etsi.org/deliver/etsi_ts/103400_103499/10341003/01.01.02_60/ts_10341003v010102p.pdf}
}

@misc{bot2021spec,
  title        = {Building Topology Ontology},
  author       = {Rasmussen, Mads Holten and Pauwels, Pieter and Lefran{\c{c}}ois, Maxime and Schneider, Georg Ferdinand},
  howpublished = {W3C Linked Building Data Community Group, Draft Community Group Report, ontology version 0.3.2},
  note         = {Section ``Alignment Modules''; undated},
  year         = {n.d.},
  url          = {https://w3c-lbd-cg.github.io/bot/}
}

@techreport{ogc2024geosparql,
  title       = {{OGC} {GeoSPARQL} -- A Geographic Query Language for {RDF} Data, Version 1.1},
  author      = {Car, Nicholas J. and Homburg, Timo and Perry, Matthew and Knibbe, Frans and Cox, Simon J. D. and Abhayaratna, Joseph and Bonduel, Mathias and Cripps, Paul J. and Janowicz, Krzysztof},
  institution = {Open Geospatial Consortium},
  type        = {OGC Standard},
  number      = {22-047r1},
  year        = {2024},
  url         = {http://www.opengis.net/doc/IS/geosparql/1.1}
}

@misc{mhclg_epc_data,
  author       = {{Ministry of Housing, Communities and Local Government}},
  title        = {Get Energy Performance of Buildings Data},
  howpublished = {GOV.UK service},
  note         = {Non-address fields under the Open Government Licence v3.0; accessed 4 November 2026},
  year         = {2026},
  url          = {https://get-energy-performance-data.communities.gov.uk/}
}

\end{document}